\documentclass[runningheads,a4paper]{llncs}

\usepackage{amssymb}
\usepackage{graphicx}
\usepackage{algorithm}
\usepackage{algpseudocode}
\usepackage{color} 

\usepackage{url}
\urldef{\mailsa}\path|{ss1, abhinavg, efros}@cs.cmu.edu|

\begin{document}

\mainmatter  

\title{Unsupervised Discovery of Mid-Level Discriminative Patches}

\titlerunning{Unsupervised Discovery of Mid-Level Discriminative Patches}

%
%
\author{Saurabh Singh \and Abhinav Gupta \and Alexei A. Efros}
\authorrunning{Singh et al.}

\institute{Carnegie Mellon University, Pittsburgh, PA 15213, USA\\
\url{http://graphics.cs.cmu.edu/projects/discriminativePatches/}}

%
%

\toctitle{Lecture Notes in Computer Science}
\tocauthor{Singh et al.}
\maketitle

\begin{abstract}
The goal of this paper is to discover a set of {\em discriminative
  patches} which can serve as a fully unsupervised
mid-level visual representation.  The desired patches need to satisfy
two requirements: 1) to be representative, they need to occur
frequently enough in the visual world; 2) to be discriminative, they
need to be different enough from the rest of the visual world.  The
patches could correspond to parts, objects, ``visual phrases'', etc.
but are not restricted to be any one of them.  We pose this
as an unsupervised discriminative
clustering problem on a huge dataset of image patches.  
We use an
iterative procedure which alternates between clustering and training
discriminative classifiers, while applying careful cross-validation at
each step to prevent overfitting. The paper experimentally
demonstrates the effectiveness
of discriminative patches as an unsupervised mid-level visual
representation, suggesting that it could be used in place of visual
words for many tasks.  Furthermore, discriminative patches can also
be used in a supervised regime, such as scene classification, where
they demonstrate state-of-the-art performance on the MIT Indoor-67
dataset.
\end{abstract}

\section{Introduction}

\begin{figure}[t]
\begin{center}
  \includegraphics[width=\textwidth]{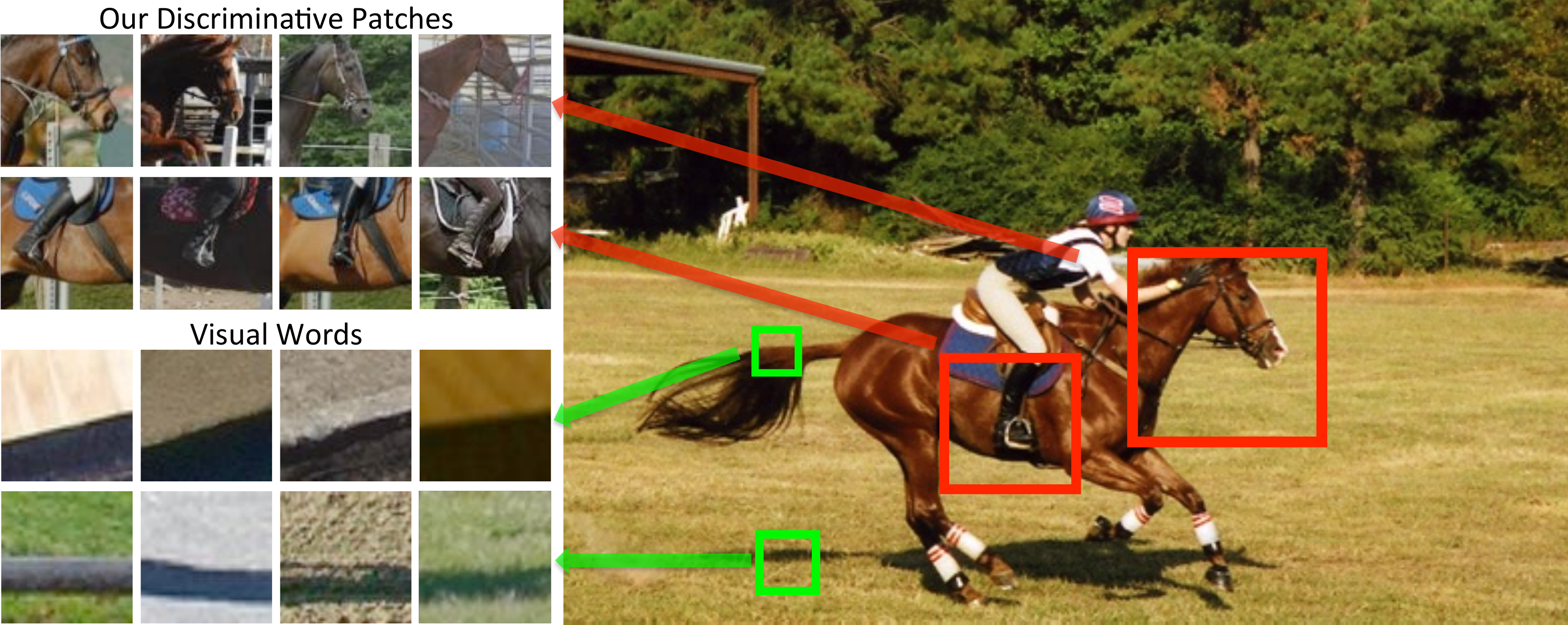}
\end{center}
   \vspace{-.25in}
   \caption{The top two detected Visual Words (bottom) vs. Mid-level
     Discriminative Patches (top), trained without any supervision and on
     the same large unlabeled dataset.}
\label{fig:teaser}
\end{figure}

Consider the image in Figure~\ref{fig:teaser}.  Shown in green are the
two most confident visual words~\cite{videogoogle} detected in this
image and the corresponding visual word clusters.  Shown in red are
the two most confident detections using our proposed mid-level
discriminative patches, computed on the same large, unlabeled image
dataset as the visual words {\em without any supervision}.  For most
people, the representation at the top seems instantly more intuitive
and reasonable.  In this paper, we will show that it is also simple to
compute, and offers very good discriminability, broad coverage, better
purity, and improved performance compared to visual word features.
Finally, we will also show how our approach can be used in a
supervised setting, where it demonstrates state-of-the-art performance
on scene classification, beating bag-of-words, spatial
pyramids~\cite{SpatialPyramid}, ObjectBank~\cite{objectbank}, and
scene deformable-parts models~\cite{Megha2011} on the MIT
Indoor-67 dataset~\cite{Indoor67}.  

What are the right primitives for representing visual information?
This is a question as old as the computer vision discipline itself,
and is unlikely to be settled anytime soon.  Over the years,
researchers have proposed a plethora of different visual features
spanning a wide spectrum, from very local to full-image, and from
low-level (bottom-up) to semantic (top-down).  In terms of spatial
resolution, one extreme is using the pixel itself as a primitive.
However there is generally not enough information at a pixel level to
make a useful feature -- it will fire all the time.  At
the other extreme, one can use the whole image as a primitive which,
while showing great promise in some
applications~\cite{tinyimages,im2gps}, requires extraordinarily large
amounts of training data, since one needs to represent all possible
spatial configurations of objects in the world explicitly.  As a
result, most researchers have converged on using features at an
intermediate scale: that of an image patch.

But even if we fix the resolution of the primitive, there is still a
wide range of choices to be made regarding {\em what} this primitive
aims to represent.  From the low-level, bottom-up point of view, an
image patch simply represents the appearance at that point, either
directly (with raw pixels~\cite{ulman}), or transformed into a
different representation (filterbank response
vector~\cite{leung_malik}, blurred~\cite{MOPS,gblur} or
spatially-binned~\cite{SIFT,HOG} feature, etc).  At a slightly higher
level, combining such patches together, typically by clustering and
histogramming, allows one to represent texture information
(e.g., textons~\cite{leung_malik}, dense bag-of-words~\cite{SpatialPyramid},
etc).  A bit higher still are approaches that encode image patches
only at sparse interest-points in a scale- and rotation-invariant way,
such as in SIFT matching~\cite{SIFT}.  Overall, the bottom-up
approaches work very well for most problems involving exact instance
matching, but their record for generalization, i.e.  finding {\em
   similar} instances, is more mixed.  One explanation is that at the
low-level it is very hard to know which parts of the representation
are the important ones, and which could be safely ignored.

As a result, recently some researchers have started looking at
high-level features, which are already impregnated with semantic
information needed to generalize well.  For example, a number of
papers have used full-blown object detectors, e.g.~\cite{pedro}, as
features to describe and reason about images
(e.g.~\cite{lorenzo,objectbank,todorovic11}).  Others have employed
discriminative part detectors such as poselets~\cite{poselets},
attribute detectors~\cite{ali10_discriminative}, ``visual
phrases''~\cite{phrases}, or ``stuff'' detectors~\cite{textonboost} as
features.  However, there are significant practical barriers to the
wide-spread adaptation of such top-down semantic techniques.  First,
they all require non-trivial amounts of hand-labeled training data per
each semantic entity (object, part, attribute, etc).  Second, many
semantic entities are just not discriminative enough visually to act
as good features.  For example, ``wall'' is a well-defined semantic
category (with plenty of training data available~\cite{SUNS}), but it
makes a lousy detector~\cite{SUNS} simply because walls are usually
plain and thus not easily discriminable.

\begin{figure*}[t]
\begin{center}
\includegraphics[width=\textwidth]{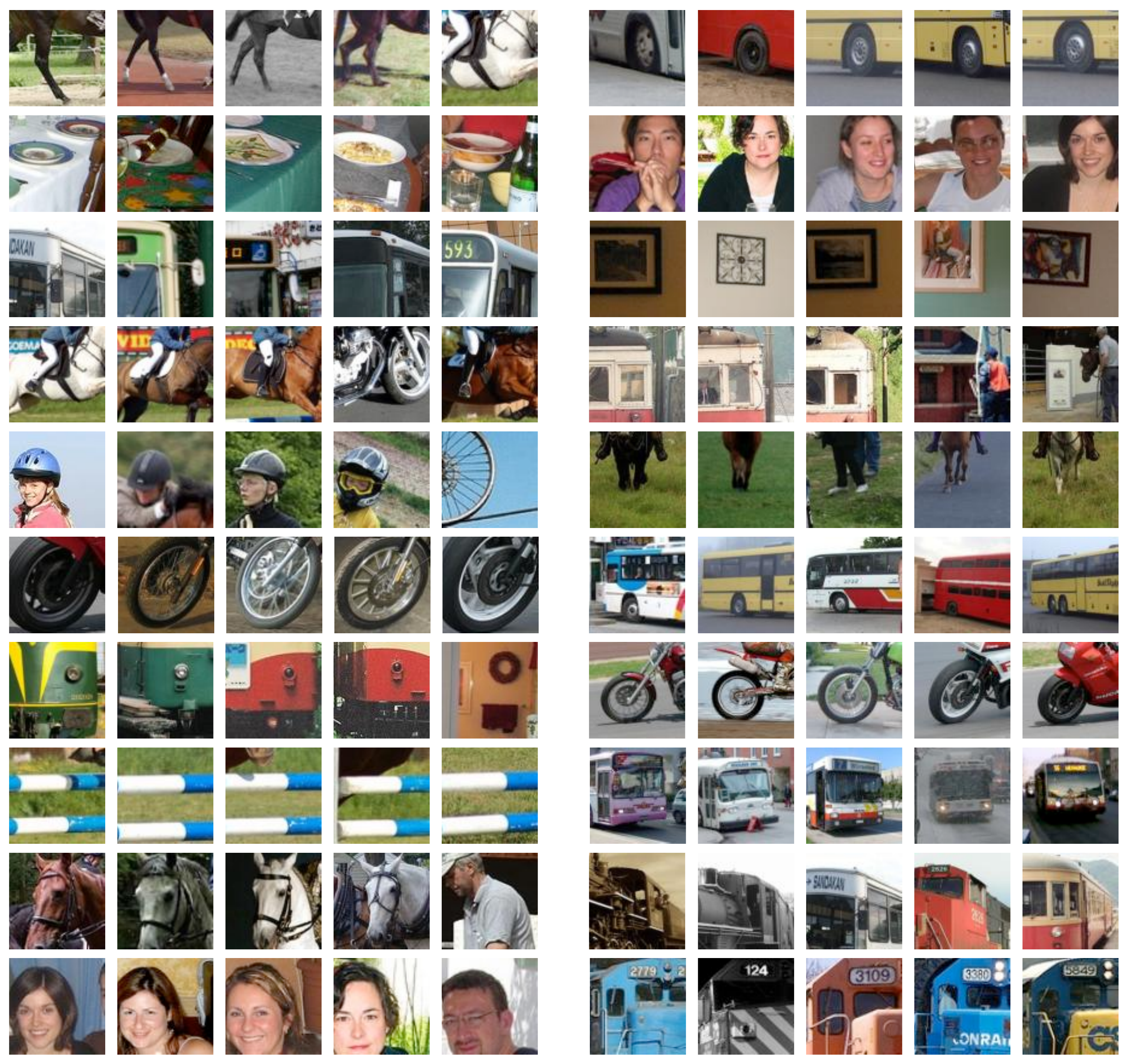}
\end{center}
\vspace{-.25in}
  \caption{Examples of discovered discriminative patches that were
    highly ranked.}
\label{fig:singlet_results}
\end{figure*}

In this paper, we consider {\em mid-level} visual primitives, which
are more adaptable to the appearance distributions in the real world
than the low-level features, but do not require the semantic grounding
of the high-level entities.  We propose a representation called {\em
  mid-level discriminative patches}.  These patches could correspond
to parts, objects, ``visual phrases'', etc. but are not restricted to
be any one of them.  What defines them is their {\em representative and discriminative
  property}: that is, that they can be detected in a large number of
images with high recall and precision.  But unlike other
discriminative methods which are weakly supervised, either with image
labels (e.g.,\cite{grouplets}) or bounding-box labels
(e.g.,~\cite{pedro}), our discriminative patches can be discovered in
a {\em fully unsupervised} manner -- given only a large pile of
unlabeled images\footnote{N.B.: The term ``unsupervised'' has changed
  its meaning over the years. E.g., while the award-winning 2003 paper
  of Fergus et al.~\cite{fergus_cvpr03} had ``unsupervised'' in its
  title, it would now be considered a weakly supervised method.}. The
key insight of this paper is to pose this as an unsupervised
discriminative clustering problem on a huge unlabeled dataset of
image patches.  We use an iterative procedure which alternates between
clustering and training discriminative classifiers (linear SVMs),
while applying careful cross-validation at each step to prevent
overfitting. Some of the
resulting discriminative patches are shown in
Figure~\ref{fig:singlet_results}.  

{\bf Prior Work:} Our goals are very much in common with prior work on finding good
mid-level feature representations, most notably the original ``visual
words'' approach~\cite{videogoogle}.  Given sparse key-point
detections over a large dataset, the idea is to cluster them in SIFT
space in an effort to yield meaningful common units of visual meaning,
akin to words in text.  However, in practice it turns out that while
some visual words do capture high-level object parts, most others
``end up encoding simple oriented bars and corners and might more
appropriately be called `visual phonemes' or even `visual
letters'.''~\cite{Russel2006}.  The way~\cite{Russel2006} addressed
these shortcomings was by using image segments as a mid-level unit for
finding commonality.  Since then, there has been a large body of work
in the general area of unsupervised object
discovery~\cite{todorovic_ahuja06,gunhee_martial08,lee_grouman08,lee_grouman10,lee_grouman11,gunhee_torralba09,henry11}.
While we share some of the same conceptual goals, our work is quite
different in that: 1) we do not explicitly aim to discover whole
semantic units like objects or parts, 2)
unlike~\cite{todorovic_ahuja06,lee_grouman08,gunhee_torralba09}, we
do not assume a single object per image, 3) whereas in object discovery
there is no separate training and test set, we explicitly aim to
discover patches that are detectable in novel images.  Because only
visual words~\cite{videogoogle} have all the above properties, that
will be our main point of comparison.

Our paper is very much inspired by poselets~\cite{poselets}, both in
its goal of finding representative yet discriminative regions, and its
use of HOG descriptors and linear SVMs.  However, poselets is a
heavily-supervised method, employing labels at the image, bounding
box, and part levels, whereas our approach aims to solve a much harder
problem without any supervision at all, so direct comparisons between
the two would not be meaningful.  Our work is also informed
by~\cite{abhinav2}, who show that discriminative machinery, such as a
linear SVM, could be successfully used in a fully unsupervised manner.

\section{Discovering Discriminative Patches}
\label{sec:singlets}

Given an arbitrary set of unlabeled images (the ``discovery
dataset'' $\mathcal{D}$), our goal is to discover a relatively small number of
{\em discriminative patches} at arbitrary resolution which can capture the
``essence'' of that data. The challenge is that the space of potential
patches (represented in this paper by HOG features~\cite{HOG}) is
extremely large since even a single image can generate tens of
thousands of patches at multiple scales.  

\subsection{Approach Motivation}

\begin{figure*}[t]
\begin{center}
\includegraphics[width=\textwidth]{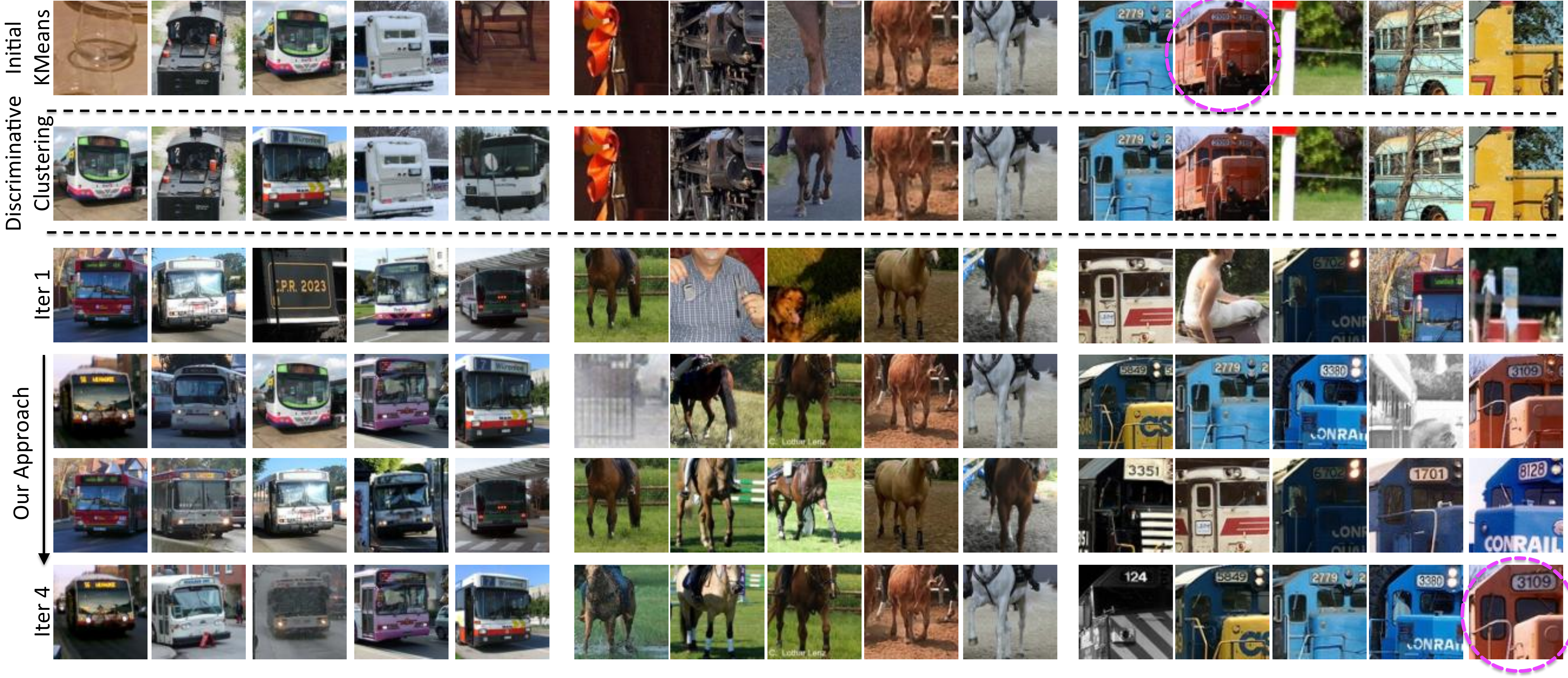}
\end{center}
\vspace{-.25in}
   \caption{Few examples to show how our iterative approach, starting
     with initial k-means clustering, converges to consistent clusters
     (Iter 4). While standard discriminative clustering approach (second
     row) also converges in some cases (1st column), in vast majority
     of cases it memorizes and overfits. Note that our
approach allows clusters to move around in x,y and scale space to find better
members (Oval in 3rd column).}

\label{fig:algo}
\end{figure*}

Of our two key requirements for good discriminative patches -- to
occur frequently, and to be sufficiently different from the rest of the visual world --
the first one is actually common to most other object discovery
approaches.  The standard solution is to employ some form of
unsupervised clustering, such as k-means, either on the entire dataset
or on a randomly sampled subset.  However, running k-means on our
mid-level patches does not produce very good clusters, as shown on
Figure~\ref{fig:algo} (Initial KMeans).  The reason is that unsupervised
clustering like k-means has no choice but to use a low-level distance
metric (e.g. Euclidean, $L1$, cross-correlation) which does not work
well for medium-sized patches, often combining instances which are in
no way visually similar.  Of course, if we somehow knew that a set of
patches {\em were} visually similar, we could easily train a
discriminative classifier, such as a linear SVM, to produce an
appropriate similarity metric for these patches.  It would seem we have
a classic chicken-and-egg problem: the clustering of the patches
depends on a good similarity, but learning a similarity
depends on obtaining good clusters.

But notice that we can pose this problem as a type of iterative
discriminative clustering.  In a typical instantiation,
e.g.~\cite{DiscKmeans}, an initial clustering of data is followed by
learning a discriminative classifier for each cluster. Based on the
discriminatively-learned similarity, new cluster memberships can be
computed by reassigning data points to each cluster, {\em etc.}.
In principle, this procedure will satisfy both of our requirements:
the clustering step will latch onto frequently occurring patches,
while the classification step will make sure that the patches in the
clusters are different enough from the rest, and thus discriminative.
However, this approach will not work on our problem ``as is'' since it is infeasible to 
use a discovery dataset large enough to be representative of the entire visual world -- it will require too many clusters.

To address this, we turn the classification step of discriminative clustering into a {\em detection} step, making each patch cluster into a detector, trained (using a linear SVM) to find
other patches like those it already owns.   This means that each cluster is now 
trained to be discriminative not just against the other clusters in
the discovery dataset $\mathcal{D}$, but against the rest of the {\em visual world},
which we propose to model by a ``natural world dataset'' $\mathcal{N}$.  The only
requirement of $\mathcal{N}$ is that it be very large (thousands of
images, containing tens of millions of patches), and drawn from a
reasonably random image distribution (we follow~\cite{abhinav2} in
simply using random photos from the Internet).    Note that $\mathcal{N}$ is not a 
``negative set'', as it can (and most likely will) contain visual patterns also found in $\mathcal{D}$ (we also experimented with $\mathcal{D} \subset \mathcal{N}$).


It is interesting to note the similarity between this version of discriminative clustering and the root filter latent updates of~\cite{pedro}.  There too, a cluster of patches (representing an object category) is being iteratively refined by making it more discriminative against millions of other image patches.  However, whereas~\cite{pedro} imposes overlap constraints preventing the cluster from moving too far from the supervised initialization, in our unsupervised formulation the clusters are completely unconstrained.


Alas, our proposed discriminative clustering procedure is still not quite enough.  Consider
Figure~\ref{fig:algo} which shows three example clusters: the top row
is simple initialization using k-means, while the second row shows the
results of the discriminative clustering described
above.  The left-most cluster shows good improvement compared to
initialization, but the other two clusters see little change. The
culprit seems to be the SVM -- it is so good at ``memorizing'' the
training data, that it is often unwilling to budge from the initial
cluster configuration. To combat this, we propose an extremely simple
but surprisingly effective solution -- cross-validation training.
Instead of training and classifying the same data, we divide our input
dataset into two equal, non-overlapping subsets.  We perform a step of
discriminative clustering on the training subset, but then apply our
learned discriminative patches on the validation subset to form
clusters there.  In this way, we are able to achieve better
generalization since the errors in the training set are largely
uncorrelated with errors in the validation set, and hence the SVM is
not able to 
overfit to them.  We then exchange the roles
of training and validation, and repeat the whole process until
convergence.  Figure~\ref{fig:algo} shows the iterations of our
algorithm for the three initial patch clusters (showing top 5 patches
in each cluster).  Note how the consistency of the clusters improves
significantly after each iteration.  Note also that the clusters can
``move around'' in $x$, $y$ and scale space to latch onto the more discriminative
parts of the visual space (see the circled train in the right-most
column).

\subsection{Approach Details}

{\bf Initialization:} The input to our discovery algorithm is a ``discovery dataset'' $\mathcal{D}$ of
unlabeled images as well as a much larger
``natural world dataset'' $\mathcal{N}$ (in this paper we used 6,000 images randomly 
sampled from Flickr.com).  First, we divide both $\mathcal{D}$ and
$\mathcal{N}$ into two equal, non-overlapping subsets ($D_1, N_1$ and $D_2, N_2$) for
cross-validation. For all images
in $\mathcal{D}_1$, we compute HOG descriptors~\cite{HOG} at
multiple resolutions (at 7 different scales).
To initialize our algorithm, we randomly sample $S$ patches from $\mathcal{D}_1$ (about
$150$ per image), disallowing highly overlapping patches or patches with
no gradient energy (e.g. sky patches) and then run standard $k$-means clustering in HOG space. 
Since we do not trust $k$-means to generalize well, we set $k$ quite high ($k=S/4$) producing tens of
thousands of clusters, most with very few members. We remove clusters
with less than $3$ patches (eliminating $66\%$ of the clusters),
ending up with about $6$ patches per image still active.

\noindent{\bf Iterative Algorithm:} 
Given an initial set of clusters $K$, we train a
linear SVM classifier~\cite{HOG} for each cluster, using
patches within the cluster as positive examples and {\em all} patches of $N_1$
as negative examples (iterative hard mining is used to handle the complexity). 
If $D_1 \subset N_1$, we exclude near-duplicates from $N_1$ by normalized 
cross-correlation $>0.4$.  The trained discriminative classifiers are then run
on the held-out validation set $D_2$, and new clusters are formed from the
top $m$ firings of each detector (we consider all SVM scores above $-1$
to be firings). We limit the new clusters to only $m=5$ members to keep
cluster purity high -- using more produces much less homogeneous
clusters.  On the other hand, if a cluster/detector fires less than 2 times
on the validation set, this suggests that it might not be very
discriminative and is killed.  The validation set now becomes the
training set and the procedure is repeated until convergence (i.e. the
top $m$ patches in a cluster do not change).  In practice, the
algorithm converges in 4-5 iterations.  The full approach is summarized in 
Algorithm~\ref{alg:algorithm1}.

\noindent{\bf Parameters:} The size of our HOG descriptor is 8x8 cells (with a
stride of 8 pixels/cell), so the minimum possible patch is 80x80
pixels, while the maximum could be as large as full image.  We use a
linear SVM (C=0.1), with 12 iterations of hard negative mining. For more details,
consult the source code on the website.

\begin{algorithm}[t]                    
\caption{Discover Top $n$ Discriminative Patches}          
\label{alg:algorithm1}
\begin{algorithmic}[1]
\Require Discovery set $\mathcal{D}$, Natural World set $\mathcal{N}$
\State $\mathcal{D} \Rightarrow \{D_1, D_2\}$;~~$\mathcal{N} \Rightarrow \{N_1, N_2\}$ \Comment{Divide $\mathcal{D,N}$ into equal sized disjoint sets}
\State $S \Leftarrow rand\_sample(D_1)$ \Comment{Sample random patches from $D_1$}
\State $K \Leftarrow kmeans(S)$ \Comment{Cluster patches using KMeans}
\While{\textbf{not} $converged()$}
\ForAll{$i$ such that $size(K[i]) \geq 3$} \Comment{Prune out small ones}
\State $C_{new}[i] \Leftarrow svm\_train(K[i], N_1)$ \Comment{Train classifier for each cluster}
\State $K_{new}[i] \Leftarrow detect\_top(C[i], D_2, m)$ \Comment{Find top m new members in other set}
\EndFor
\State $K \Leftarrow K_{new};~~ C \Leftarrow C_{new}$
\State $swap(D_1, D_2);~~ swap(N_1, N_2)$ \Comment{Swap the two sets}
\EndWhile
\State $A[i] \Leftarrow purity(K[i]) + \lambda \times discriminativeness(K[i])~\forall~i$ \Comment{Compute scores} \\
\Return $select\_top(C, A, n)$ \Comment{Sort according to scores and select top n patches}
\end{algorithmic}
\end{algorithm}

\subsection{Ranking Discriminative Patches}

Our algorithm produces a dictionary of a few thousand
discriminative patches of varying quality. Our next task is to rank
them, to find a small number of the most
discriminative ones.  Our criteria for ranking consists of two terms:

\noindent {\bf Purity:} Ideally, a good cluster should have
all its member patches come from the same visual concept.
However, measuring purity in an unsupervised setting is
impossible. Therefore, we approximate the purity of each cluster in
terms of the classifier confidence of the cluster members (assuming
that cross-validation removed overfitting). Thus, the purity score for a cluster is 
computed by summing up the SVM detection scores of top $r$ cluster
members (where $r>m$ to evaluate the generalization of the cluster
beyond the $m$ training patches).

\noindent {\bf Discriminativeness:} In an unsupervised setting, the
only thing we can say is that a highly discriminative patch should
fire {\em rarely} in the natural world.  Therefore, we define discriminativeness of a patch
as the ratio of the number of firings on $\mathcal{D}$ to
the number of firings on $\mathcal{D \cup N}$ (of course, we do not
 want patches that never fire at all, but these would
have already been removed in cross-validation training).

\begin{figure}[t]
\begin{center}
\includegraphics[width=\textwidth]{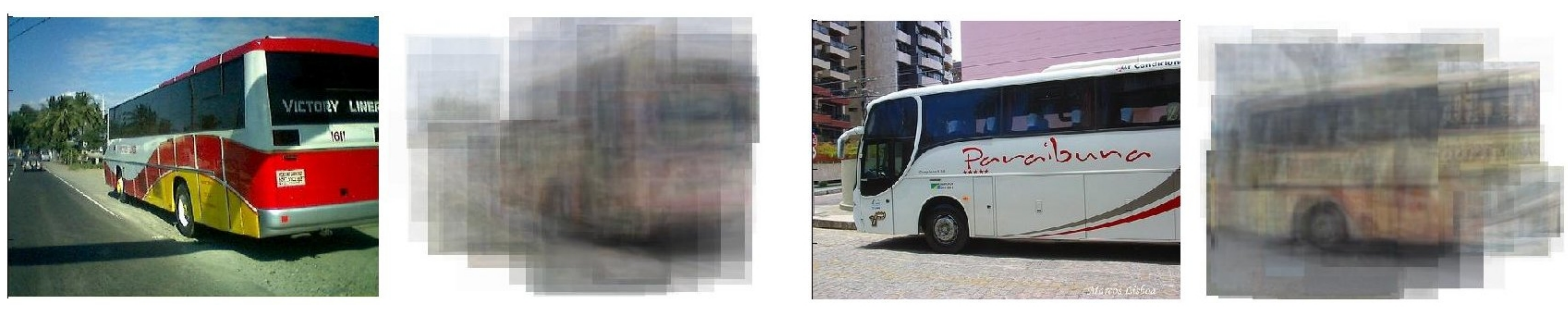}
\end{center}
\vspace{-.25in}
  \caption{Visualizing images (left) in terms of their most
    discriminative patches (right). The patch detectors were fired on
    a novel image and the high-scoring patch detections were averaged
    together, weighted by their scores.}
\label{fig:cool_buses}
\end{figure}

All clusters are ranked using a linear combination of the above two
scores. Figure~\ref{fig:singlet_results} shows a set of top-ranked
discriminative patch clusters discovered with our approach.  Note how
sometimes the patches correspond to object parts, such as ``horse
legs'' and ``horse muzzle'', sometimes to whole objects, such as
``plates'', and sometimes they are just discriminative portions of an
object, similar to poselets (e.g., see the corner of trains).  Also note
that they exhibit surprisingly good visual consistency for a fully
unsupervised approach.  The ability of discriminative patches to fire
on visually similar image regions is further demonstrated in
Figure~\ref{fig:cool_buses}, where the patch detectors are applied to
a novel image and high-scoring detections are displayed with the
average patch from that cluster.  In a way, the figure shows what our
representation captures about the image.

\section{Discovering ``Doublets''}
\label{sec:doublets}

\begin{figure}[t]
\begin{center}
\includegraphics[width=\textwidth]{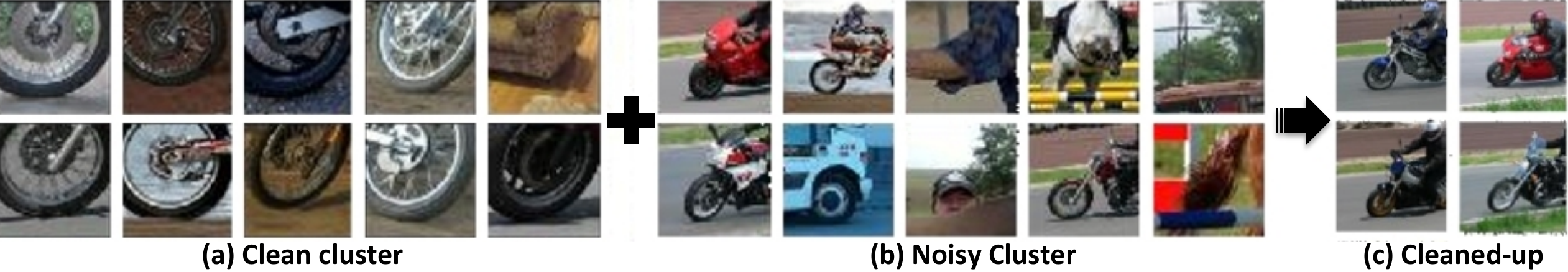}
\end{center}
\vspace{-.25in}
  \caption{Cluster clean-up using ``doublets''. A visually
    non-homogeneous cluster (b) that has learned more than one
    concept, when coupled into a doublet with a high quality
    cluster (a), gets cleaned up (c).}

\label{fig:two_concepts}
\end{figure}

While our discriminative patch discovery approach is able to produce a
number of visually good, highly-ranked discriminative patches, some
other potentially promising ones do not make it to the top due to low
purity.  This happens when a cluster converges to two or more
``concepts'' because the underlying classifier is able to generalize
to both concepts simultaneously
(e.g., Figure~\ref{fig:two_concepts}b). However, often the two concepts
have different firing patterns with respect to some other mid-level
patch in the dictionary, e.g., motorcycle wheel in
Figure~\ref{fig:two_concepts}a.  Therefore, we propose to employ second-order
spatial co-occurrence relationships among our
discriminative patches as a way of ``cleaning them up''
(Figure~\ref{fig:two_concepts}c).  Moreover, discovering these
second-order relationships can provide us with
``doublets''~\cite{sivic05} (which could be further generalized to
grouplets~\cite{grouplets,UFO}) that can themselves be highly
discriminative and useful as mid-level features in their own right.

\begin{figure*}[t]
\begin{center}
\includegraphics[width=\textwidth]{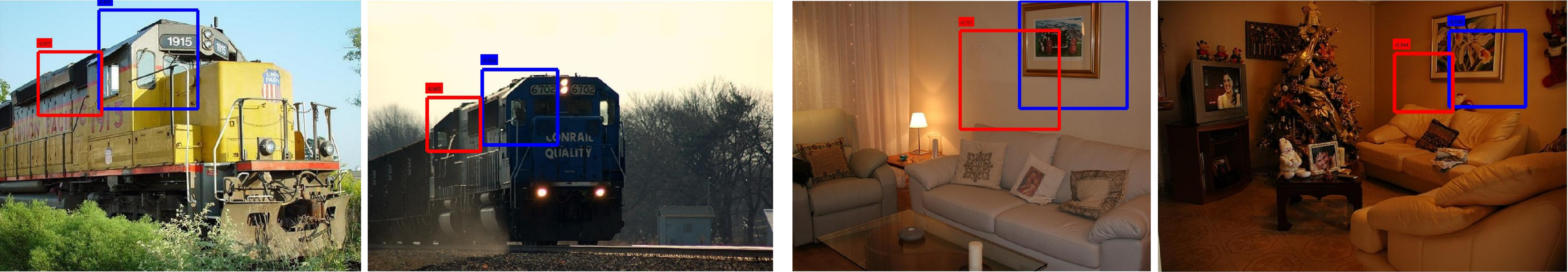}
\end{center}
\vspace{-.25in}
  \caption{Examples of discovered discriminative ``doublets'' that
    were highly ranked.}

\label{fig:doublets}
\end{figure*}

To discover doublets, we start with a list of highly discriminative
patches that will serve as high-quality ``roots''.  For each root
patch, we search over all the other discovered discriminative patches
(even poor-quality ones), and record their relative spatial
configuration in each image where they both
fire.  The pairs that exhibit a highly spatially-correlated firing
pattern become potential doublets.  We rank the doublets by applying
them on the (unlabeled) validation set.  The doublets are ranked high
if in images where both patches fire, their relative spatial
configuration is consistent with what was observed in the training
set. In Figure~\ref{fig:doublets} we show some examples of highly
discriminative doublets. Notice that not only is the quality of
discriminative patches good, but also the spatial relationships
within the doublet are intuitive. 


\section{Quantitative Evaluation}
\label{sec:results}

As with other unsupervised discovery approaches, evaluation is
difficult.  We have shown a number of qualitative results
(Figures~\ref{fig:singlet_results},~\ref{fig:doublets}), and there are
many more on the website.  For the first set of
quantitative evaluations (as well as for all the qualitative results
except Figure~\ref{fig:indoorSceneResults}), we have chosen a subset
of of PASCAL VOC 2007~\cite{pascal} (1,500 images) as our discovery
dataset.  We picked PASCAL VOC because it is a well-known and
difficult dataset, with rich visual diversity and scene clutter.
Moreover, it provides annotations for a number of object classes
which could be used to evaluate our unsupervised discovery.  However,
since our discovered patches are not meant to correspond to semantic
objects, this evaluation metric should be taken with quite a few
grains of salt.

One way to evaluate the quality of our discriminative patch clusters
is by using the standard unsupervised discovery measures of ``purity''
and ``coverage'' (e.g.,~\cite{henry11}).  {\em Purity} is defined by
what percentage of cluster members correspond to the same visual
entity.  In our case, we will use PASCAL semantic category annotations
as a surrogate for visual similarity.  For each of the top $1000$
discovered patches, we first assign it to one of the semantic
categories using majority membership.  We then measure purity as
percentage of patches assigned to the same PASCAL semantic label. {\em
  Coverage} is defined as the number of images in the dataset
``covered'' (fired on) by a given cluster.  

\begin{figure}[t]
\begin{center}
\includegraphics[width=0.48\textwidth]{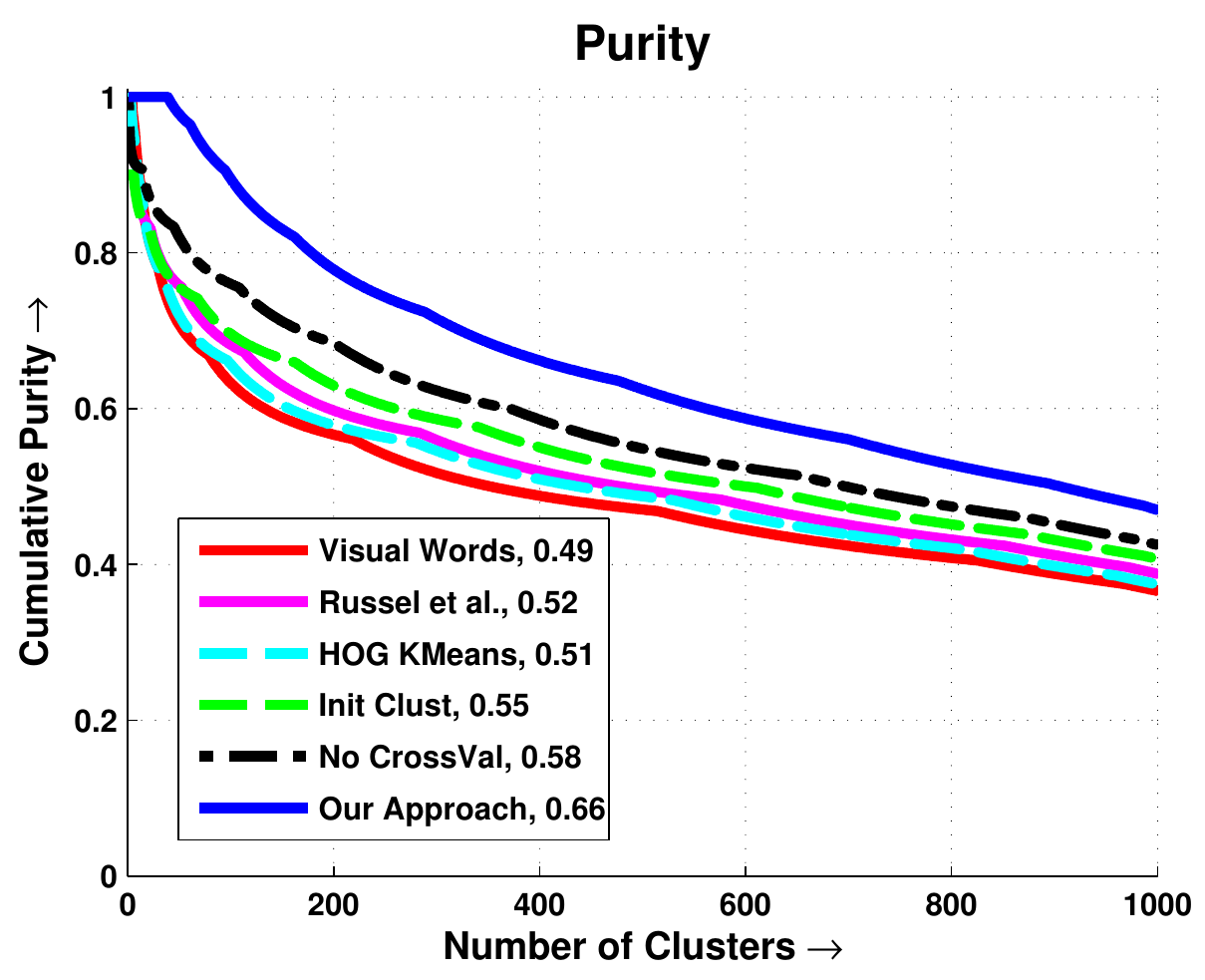}
\includegraphics[width=0.48\textwidth]{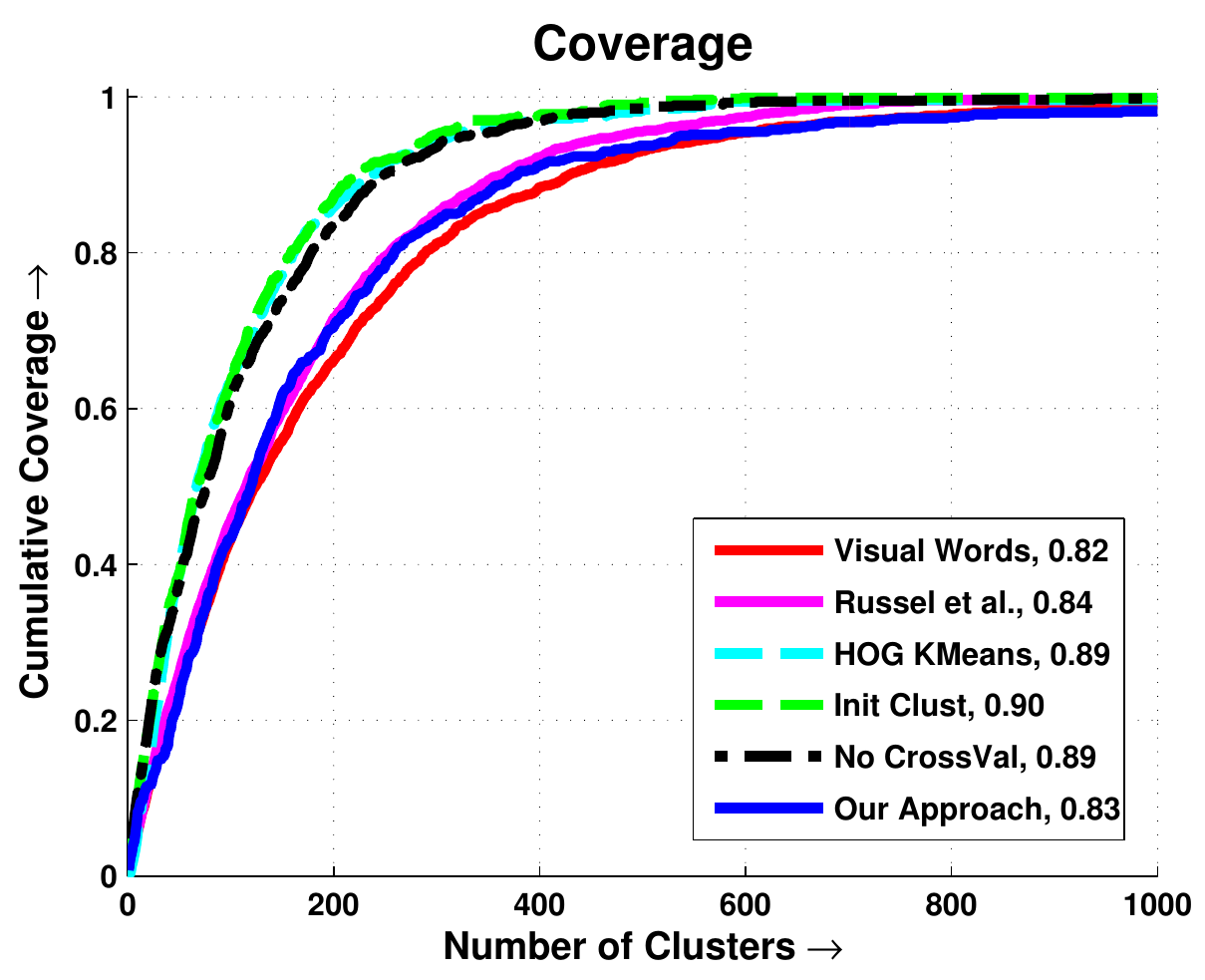}
\end{center}
\vspace{-.25in}
  \caption{Quantitative comparison of discriminative patches compared
    to the baseline approaches. Quality of clustering is evaluated in
    terms of the area under the curve for cumulative purity and
    coverage.}

\label{fig:graphs}
\end{figure}

Figure~\ref{fig:graphs} reports the purity and coverage of our
approach and a number of baselines. For each one, the graphs show
the cumulative purity/coverage as number of clusters being considered
is increased (the clusters are sorted in the decreasing order of
purity). We compare our approach with Visual Words~\cite{videogoogle}
and Russell et. al~\cite{Russel2006} baseline, plus a number of
intermediate results of our method: 1) HOG K-Means (visual word analog
for HOG features), 2) Initial Clustering (SVMs trained on the K-Means
clusters without discriminative re-clustering), and 3) No
Cross-Validation (iterative, discriminatively-trained clusters but
without cross-validation).  In each case, the numbers indicate
area-under-the-curve (AUC) for each method.  Overall, our approach
demonstrates substantial gain in purity without sacrificing much
coverage as compared to the established approaches.  Moreover, each
step of our algorithm improves purity.  Note in particular the
substantial improvement afforded by the cross-validation
training procedure compared to standard training.

As we mentioned, however, the experiment above under-reports the
purity of our clusters, since semantic equivalence is not the same as
visual similarity.  Therefore, we performed an informal perceptual
experiment with human subjects, measuring the {\em visual purity} of
our clusters.  We selected the top 30 clusters from the dataset. For
each cluster, we asked human labelers to mark which of the cluster's
top ten firings on the validation set are visually consistent with the
cluster.  Based on this measure, average visual purity for these
clusters was 73\%.

\begin{table*}[!t]
\centering
\begin{tabular}{|l||c|}
\hline
GIST & 29.7 \\
Spatial Pyramid HOG $(SPHOG)^{\dagger}$ & 29.8\\
Spatial Pyramid SIFT $(SP)^{\dagger}$ & 34.4 \\
ROI-GIST~\cite{Indoor67} & 26.5 \\
Scene DPM~\cite{Megha2011} & 30.4 \\
MM-Scene~\cite{Zhu_LargeMargin} & 28.0 \\
Object Bank~\cite{objectbank} & 37.6 \\
\hline
Ours & \bf{38.1} \\
\hline
\end{tabular}
\quad
\begin{tabular}{|l||c|}
\hline

Ours+GIST & 44.0 \\
Ours+SP & 46.4 \\
Ours+GIST + SP & 47.5 \\
Ours+DPM & 42.4 \\
Ours+GIST+DPM & 46.9 \\
Ours+SP+DPM & 46.4 \\
\hline
GIST+SP+DPM~\cite{Megha2011} & 43.1* \\
Ours+GIST+SP+DPM & {\bf 49.4} \\
\hline
\end{tabular}
\vspace{.05In}
\caption{Quantitative Evaluation: Average Classification on MIT Indoor-67 dataset.
 *Current state-of-the-art. $\dagger$Best performance from
  various vocabulary sizes.}

\label{indoorTable}
\end{table*}



\subsection{Supervised Image Classification}
\label{sec:classification}

Unsupervised clustering approaches, such as visual words, have long
been used as features for supervised tasks, such as classification.
In particular, bag of visual words and spatial
pyramids~\cite{SpatialPyramid} are some of the most popular current
methods for image classification.  Since our mid-level patches
could be considered the true visual words (as opposed to ``visual
letters''), it makes sense to see how they would perform on a
supervised classification task.  We evaluate them in two different
settings: 1) unsupervised discovery, supervised classification, and 2)
supervised discovery, supervised classification.

\begin{figure*}[!t]
\begin{center}
\includegraphics[width=\textwidth]{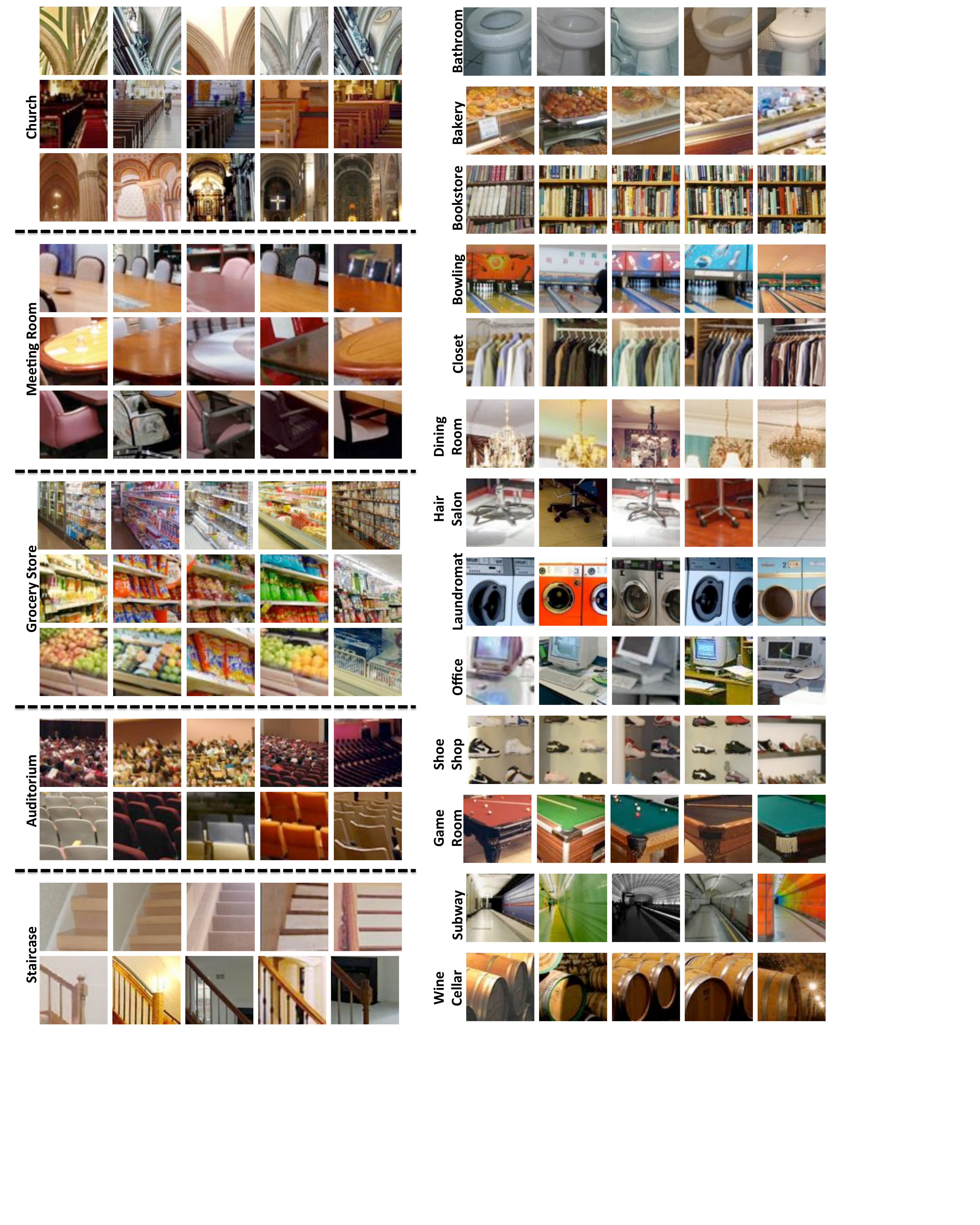}
\end{center}
\vspace{-.25in}
  \caption{Top discriminative patches for a sampling of scenes in
    the MIT Indoor-67 Scene dataset~\cite{Indoor67}. Note how these
    capture various visual aspects of a typical scene.}
\label{fig:indoorSceneResults}
\end{figure*}

\noindent
{\bf Unsupervised discriminative patches} \\
Using the discriminative patches discovered from the same PASCAL VOC
discovery dataset as before, we would like to see if they could make
better visual words for a supervised image classification task.  Our
baseline is the standard spatial pyramid of visual words (using $1000$
visual words) using their public code~\cite{SpatialPyramid}.  
For our approach, we construct spatial pyramid using top 1000
discriminative patches.
Classification was performed using a simple linear SVM and performance
was evaluated using Average Precision. Standard
visual words scored $0.54$ AP, while using our discriminative patches,
the score was $0.65$ AP.  We further expanded our feature
representation by adding the top $250$-ranking doublets as extra
visual words, resulting in a slight improvement to $0.66$ AP.

\noindent
{\bf Supervised discriminative patches} \\
We further want to evaluate the performance of our approach when it is
allowed to utilize more supervision for a fair comparison with several
existing supervised approaches. Instead of discovering the
discriminative patches from a common pool of all the images, we can
also discover them on a per-category basis.  In this experiment, we
perform supervised scene classification using the challenging MIT
Indoor-67 dataset~\cite{Indoor67}, containing 67 scene categories.
Using the provided scene labels, we discover discriminative patches
for each scene independently, while treating all other images in the
dataset as the ``natural world".  

Figure~\ref{fig:indoorSceneResults} shows top few most discriminative
patches discovered this way for a number of categories from the
dataset. It is interesting to see that the discriminative patches
capture aspects of scenes that seem very intuitive to us. In
particular the discriminative patches for the \emph{Church} category
capture the arches and the benches; the ones for the \emph{Meeting
  Room} capture the center table and the seats. These discriminative
patches are therefore capturing the essence of the scene in terms of
these highly consistent and repeating patterns and hence providing a
simple yet highly effective mid-level representation. Inspired by these results, we have also applied a similar approach to discovering ``What makes Paris look like Paris"~\cite{doersch2012what} 
using geographic labels as the weak supervisory signal. 

To perform classification, top 210 discovered patches of each scene are aggregated into a
spatial pyramid using maxpooling over the discriminative patch scores
as in~\cite{objectbank}. We again use a linear SVM in a
one-vs-all classification. The results are reported in
Table~\ref{indoorTable}. Comparison with HOG visual words (SPHOG) shows the huge performance gain resulting from our algorithm 
when operating in the same feature space. Further, our simple method by itself
outperforms all others that have been tested on this
dataset~\cite{Indoor67,Megha2011,Zhu_LargeMargin,objectbank}.
Moreover, combining our method with the currently best-performing
combination approach of~\cite{Megha2011} yields {\bf49.4}\%
performance which, to our knowledge, is the best on this dataset.

{
\noindent
{\bf\\ Acknowledgments:} The authors would like to thank Martial Hebert, Tomasz Malisiewicz,
Abhinav Shrivastava and Carl Doersch for many helpful discussions. This work was 
supported by ONR Grant N000141010766.
}

\bibliographystyle{splncs}
\bibliography{egbib}

\begin{thebibliography}{10}

\bibitem{videogoogle}
Sivic, J., Zisserman, A.:
\newblock {Video Google}: {A} text retrieval approach to object matching in
  videos.
\newblock In: ICCV. (2003)

\bibitem{SpatialPyramid}
Lazebnik, S., Schmid, C., Ponce, J.:
\newblock Beyond bags of features: Spatial pyramid matching for recognizing
  natural scene categories.
\newblock In: CVPR. (2006)

\bibitem{objectbank}
Li, L.J., Su, H., Xing, E., Fei-fei, L.:
\newblock Object bank: A high-level image representation for scene
  classification and semantic feature sparsification,.
\newblock NIPS (2010)

\bibitem{Megha2011}
Pandey, M., Lazebnik, S.:
\newblock Scene recognition and weakly supervised object localization with
  deformable part-based models.
\newblock In: ICCV. (2011)

\bibitem{Indoor67}
Quattoni, A., Torralba, A.:
\newblock Recognizing indoor scenes.
\newblock In: CVPR. (2009)

\bibitem{tinyimages}
Torralba, A., Fergus, R., Freeman, W.T.:
\newblock 80 million tiny images: a large database for non-parametric object
  and scene recognition.
\newblock PAMI (2008)

\bibitem{im2gps}
Hays, J., Efros, A.A.:
\newblock im2gps: estimating geographic information from a single image.
\newblock In: CVPR. (2008)

\bibitem{ulman}
Ullman, Vidal-Naquet, S., E., M.S.:
\newblock Visual features of intermediate complexity and their use in
  classification.
\newblock NATURE AMERICA (2002)

\bibitem{leung_malik}
Leung, T., Malik, J.:
\newblock Representing and recognizing the visual appearance of materials using
  three-dimensional textons.
\newblock (2001)

\bibitem{MOPS}
Brown, M., Szeliski, R., Winder, S.:
\newblock Multi-image matching using multi-scale oriented patches.
\newblock In: CVPR. (2005)

\bibitem{gblur}
Berg, A., Malik, J.:
\newblock Geometric blur for template matching.
\newblock In: CVPR. (2001)

\bibitem{SIFT}
Lowe, D.:
\newblock Distinctive image features from scale-invariant keypoints.
\newblock IJCV (2004)

\bibitem{HOG}
Dalal, N., Triggs, B.:
\newblock Histograms of oriented gradients for human detection.
\newblock In: CVPR. (2005)

\bibitem{pedro}
Felzenszwalb, P., McAllester, D., Ramanan, D.:
\newblock A discriminatively trained, multiscale, deformable part model.
\newblock CVPR (2008)

\bibitem{lorenzo}
Torresani, L., Szummer, M., Fitzgibbon, A.:
\newblock Efficient object category recognition using classemes.
\newblock In: ECCV. (2010)

\bibitem{todorovic11}
Payet, N., Todorovic, S.:
\newblock Scene shape from texture of objects.
\newblock In: CVPR. (2011)

\bibitem{poselets}
Bourdev, L., Malik, J.:
\newblock Poselets: Body part detectors trained using 3d human pose
  annotations.
\newblock In: ICCV. (2009)

\bibitem{ali10_discriminative}
Farhadi, A., Endres, I., Hoiem, D.:
\newblock Attribute-centric recognition for cross-category generalization.
\newblock In: CVPR. (2010)

\bibitem{phrases}
Sadeghi, M., Farhadi, A.:
\newblock Recognition using visual phrases.
\newblock In: CVPR. (2011)

\bibitem{textonboost}
Shotton, J., Winn, J., Rother, C., Criminsi, A.:
\newblock Textonboost: Joint appearance, shape and context modeling for
  multi-class object recognition and segmentation.
\newblock In: ECCV. (2006)

\bibitem{SUNS}
Choi, M.J., Lim, J.J., Torralba, A., Willsky, A.S.:
\newblock Exploiting hierarchical context on a large database of object
  categories.
\newblock In: CVPR. (2010)

\bibitem{grouplets}
Yao, B., Fei-Fei, L.:
\newblock Grouplet: A structured image representation for recognizing human and
  object interactions.
\newblock In: CVPR. (2010)

\bibitem{fergus_cvpr03}
Fergus, R., Perona, P., Zisserman, A.:
\newblock Object class recognition by unsupervised scale-invariant learning.
\newblock In: CVPR. (2003)

\bibitem{Russel2006}
Russell, B., Freeman, W., Efros, A., Sivic, J., Zisserman, A.:
\newblock Using multiple segmentations to discover objects and their extent in
  image collections.
\newblock In: CVPR. (2006)

\bibitem{todorovic_ahuja06}
Todorovic, S., Ahuja, N.:
\newblock Unsupervised category modeling, recognition, and segmentation in
  images.
\newblock PAMI (2008)

\bibitem{gunhee_martial08}
Kim, G., Faloutsos, C., Hebert, M.:
\newblock {Unsupervised Modeling of Object Categories Using Link Analysis
  Techniques}.
\newblock In: CVPR. (2008)

\bibitem{lee_grouman08}
Lee, Y.J., Grauman, K.:
\newblock Foreground focus: Unsupervised learning from partially matching
  images.
\newblock IJCV (2009)

\bibitem{lee_grouman10}
Lee, Y.J., Grauman, K.:
\newblock Object-graphs for context-aware category discovery.
\newblock CVPR (2010)

\bibitem{lee_grouman11}
Lee, Y.J., Grauman, K.:
\newblock Learning the easy things first: Self-paced visual category discovery.
\newblock In: CVPR. (2011)

\bibitem{gunhee_torralba09}
Kim, G., Torralba, A.:
\newblock {Unsupervised Detection of Regions of Interest using Iterative Link
  Analysis}.
\newblock In: NIPS. (2009)

\bibitem{henry11}
Kang, H., Hebert, M., Kanade, T.:
\newblock Discovering object instances from scenes of daily living.
\newblock In: ICCV. (2011)

\bibitem{abhinav2}
Shrivastava, A., Malisiewicz, T., Gupta, A., Efros, A.A.:
\newblock Data-driven visual similarity for cross-domain image matching.
\newblock ACM ToG (SIGGRAPH Asia) (2011)

\bibitem{DiscKmeans}
Ye, J., Zhao, Z., Wu, M.:
\newblock Discriminative k-means for clustering.
\newblock In: NIPS. (2007)

\bibitem{sivic05}
Sivic, J., Russell, B.C., Efros, A.A., Zisserman, A., Freeman, W.T.:
\newblock Discovering object categories in image collections.
\newblock In: ICCV. (2005)

\bibitem{UFO}
Karlinsky, L., Dinerstein, M., Ullman, S.:
\newblock Unsupervised feature optimization (ufo): Simultaneous selection of
  multiple features with their detection parameters.
\newblock In: CVPR. (2009)

\bibitem{pascal}
Everingham, M., Gool, L.V., Williams, C.K.I., Winn, J., Zisserman, A.:
\newblock The {PASCAL} {V}isual {O}bject {C}lasses {C}hallenge (2007)

\bibitem{Zhu_LargeMargin}
Zhu, J., Li, L.J., Li, F.F., Xing, E.P.:
\newblock Large margin learning of upstream scene understanding models.
\newblock In: NIPS. (2010)

\bibitem{doersch2012what}
Doersch, C., Singh, S., Gupta, A., Sivic, J., Efros, A.A.:
\newblock What makes paris look like paris?
\newblock ACM Transactions on Graphics (SIGGRAPH) \textbf{31} (2012)

\end{thebibliography}

\end{document}